\documentclass[10pt,conference]{IEEEtran}
\IEEEoverridecommandlockouts
\usepackage{cite}
\usepackage{amsmath,amssymb,amsfonts}
\usepackage{algorithmic}
\usepackage{graphicx}
\usepackage{textcomp}
\usepackage{xcolor}
\usepackage{float}
\usepackage{multirow} % For multi-row cells
\usepackage{array}    % For defining column styles
\usepackage{booktabs} % For better horizontal lines
\usepackage{afterpage}
\usepackage[caption=false,font=footnotesize]{subfig}
\usepackage{array} % Add this for m{width} functionality
\usepackage[hidelinks]{hyperref}

\hypersetup{
    colorlinks=true,
    linkcolor=blue,      % or black for print
    citecolor=green,     % dark green or black for a more subdued look
    urlcolor=blue,       % dark red or a different shade of blue for distinction
    anchorcolor=blue     % color for anchor text
}

\makeatletter
  \ifCLASSOPTIONconference
    \long\def\@makecaption#1#2{%
      \ifx\@captype\@IEEEtablestring%
        \footnotesize
        \bgroup\par\centering\@IEEEtabletopskipstrut{\normalfont\footnotesize {#1.} #2}\par\addvspace{0.5\baselineskip}\egroup%
        \@IEEEtablecaptionsepspace
      \else
        \@IEEEfigurecaptionsepspace
        \setbox\@tempboxa\hbox{\normalfont\footnotesize {#1.} #2}%
        \ifdim \wd\@tempboxa >\hsize%
          \setbox\@tempboxa\hbox{\normalfont\footnotesize {#1.}}%
          \parbox[t]{\hsize}{\normalfont\footnotesize \noindent\unhbox\@tempboxa#2}%
        \else
          \hbox to\hsize{\normalfont\footnotesize\hfil\box\@tempboxa\hfil}%
        \fi
      \fi}
  \fi
\makeatother

\def\BibTeX{{\rm B\kern-.05em{\sc i\kern-.025em b}\kern-.08em
    T\kern-.1667em\lower.7ex\hbox{E}\kern-.125emX}}
\begin{document}

% \title{VQAs in Computer Vision: NISQ-Seg Image Segmentation}
\title{Qubit-efficient Variational Quantum Algorithms for Image Segmentation}

\author{
    Supreeth Mysore Venkatesh\textsuperscript{1,2},
    Antonio Macaluso\textsuperscript{2},
    Marlon Nuske\textsuperscript{3},\\
    Matthias Klusch\textsuperscript{2}, 
    Andreas Dengel\textsuperscript{1,3}\\
    \textsuperscript{1}\textit{University of Kaiserslautern-Landau (RPTU)}, Kaiserslautern, Germany \\
    \textsuperscript{2}\textit{German Research Center for Artificial Intelligence (DFKI)}, Saarbruecken, Germany \\
    \textsuperscript{3}\textit{German Research Center for Artificial Intelligence (DFKI)}, Kaiserslautern, Germany \\
    % \href{mailto:supreeth.mysore@rptu.de}
    {\texttt{supreeth.mysore@rptu.de}}, 
    % \href{mailto:antonio.macaluso@dfki.de}
    {\texttt{antonio.macaluso@dfki.de}}, 
    % \href{mailto:marlon.nuske@dfki.de}
    {\texttt{marlon.nuske@dfki.de}},\\
    % \href{mailto:matthias.klusch@dfki.de}
    {\texttt{matthias.klusch@dfki.de}}, 
    % \href{mailto:andreas.dengel@dfki.de}
    {\texttt{andreas.dengel@dfki.de}}
}

\maketitle

\vspace*{-30pt}

\begin{abstract}

Quantum computing is expected to transform a range of computational tasks beyond the reach of classical algorithms.
In this work, we examine the application of variational quantum algorithms (VQAs) for unsupervised image segmentation to partition images into separate semantic regions.
Specifically, we formulate the task as a graph cut optimization problem and employ two established qubit-efficient VQAs, which we refer to as Parametric Gate Encoding (PGE) and Ancilla Basis Encoding (ABE), to find the optimal segmentation mask.
In addition, we propose Adaptive Cost Encoding (ACE), a new approach that leverages the same circuit architecture as ABE but adopts a problem-dependent cost function. 
We benchmark PGE, ABE and ACE on synthetically generated images, focusing on quality and trainability.
ACE shows consistently faster convergence in training the parameterized quantum circuits in comparison to PGE and ABE.
Furthermore, we provide a theoretical analysis of the scalability of these approaches against the Quantum Approximate Optimization Algorithm (QAOA), showing a significant cutback in the quantum resources, especially in the number of qubits that logarithmically depends on the number of pixels.
The results validate the strengths of ACE, while concurrently highlighting its inherent limitations and challenges.
This paves way for further research in quantum-enhanced computer vision.

\end{abstract}


\vspace*{-6pt}

\section{Introduction}

\vspace*{-4pt}
% The pursuit of quantum advantage is driving the evolution of computing technologies, aiming to unlock breakthroughs in various applications such as prime number factorization, database searching, quantum system simulations, and solving linear equations. The introduction of cloud-based quantum computing marked a significant milestone, enabling access to Noisy Intermediate-Scale Quantum (NISQ) devices for researchers worldwide. However, the use of NISQ devices is challenged by noise and limited qubit count, hindering the full potential of quantum algorithms. Despite these obstacles, research is intensively directed toward identifying computational tasks that could benefit from the current capabilities of NISQ devices. Fields such as artificial intelligence, computer vision, financial modeling, computational chemistry, and drug development are key areas of focus, as they involve processing vast amounts of data and complex computations that NISQ devices might effectively tackle.
% The pursuit of quantum advantage aims to revolutionize areas like prime factorization and database searching through cloud-based quantum computing and NISQ devices. Despite challenges like noise and qubit limitations, research targets utilizing NISQ for fields requiring intensive data processing and complex computations, such as artificial intelligence and drug development.
Variational Quantum Algorithms (VQAs) aim to leverage NISQ devices for practical uses through a hybrid quantum-classical approach \cite{Cerezo2021}. 
VQAs address classical problems, which lack polynomial-time solutions but have verifiable answers, showcasing their potential for real-world applications.
One application area that highlights the potential of VQAs is in solving Quadratic Unconstrained Binary Optimization (QUBO) problems, an NP-Hard class of optimization problems with applications in various domains, including logistics, finance, and machine learning \cite{glover2022applications}.
The Quantum Approximate Optimization Algorithm (QAOA) represents a notable example of VQAs applied to such challenges \cite{farhi2014quantum}.
% Although solving QUBO, an NP-hard problem, is feasible on classical computers for instances with a few thousand variables, executing QAOA to address similar tasks demands thousands of qubits—a requirement that appears out of reach given the constraints of near-term quantum technology.
While classical computers can handle QUBO for instances with several thousand variables, running QAOA on similar tasks requires thousands of qubits, a requirement that appears out of reach given the constraints of near-term quantum technology.
The scalability of QAOA is inherently limited by the linear relationship between the number of binary variables in QUBO problems and the number of logical qubits required, underscoring the pressing need to explore more efficient encoding strategies. 
% These strategies aim not only to alleviate the hardware constraints of current quantum devices but also to unlock new possibilities for efficiently solving specific complex tasks that remain beyond the reach of classical computing capabilities.
These strategies are designed to mitigate the hardware limitations of current quantum devices while simultaneously unlocking new possibilities for efficiently addressing complex tasks that classical computing capabilities cannot solve.

\begin{figure}[t]
\centering
\centerline{\includegraphics[width=\columnwidth]{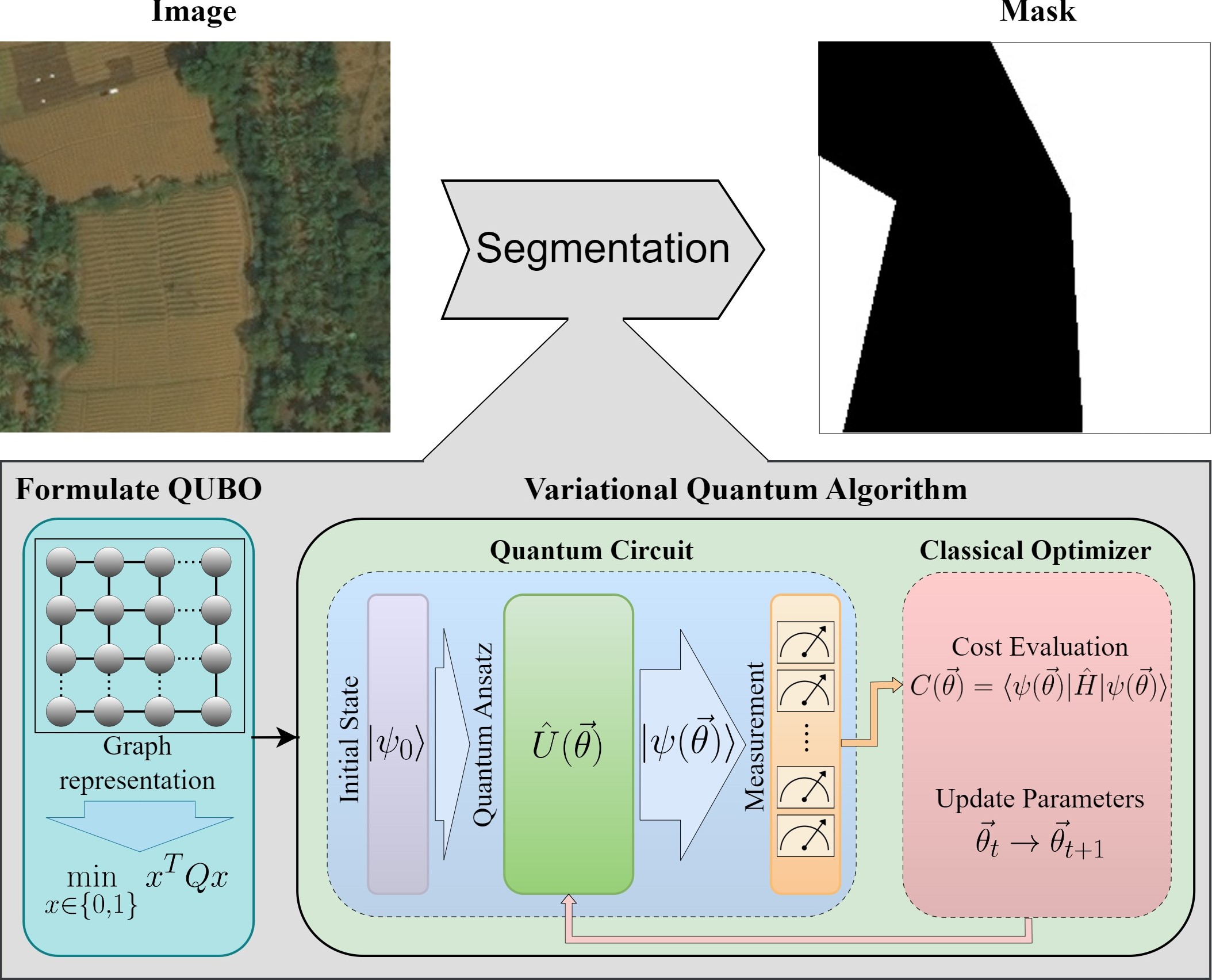}}
\caption{The figure depicts an architecture for segmenting an image. Generating a graph representation of the image, the distinct semantic regions in the image are identified by finding the minimum cut in the graph that can be formulated as Quadratic Unconstrained Binary Optimization (QUBO). The QUBO problem is subsequently solved using a VQA by optimizing the circuit parameters that incidentally explore an exponentially large solution space to locate the global optimum, which can be decoded as the segmentation mask.}
\vspace*{-18pt}
\label{fig: poster}
\end{figure}

% Variational Quantum Algorithms (VQAs) stand at the forefront of this exploration, offering a promising avenue to harness the power of NISQ devices for practical applications. By adopting a hybrid quantum-classical approach, VQAs tackle NP-hard problems—tasks for which no polynomial-time algorithm exists for their exact solution in the classical realm, yet their solutions are relatively straightforward to verify. 

In this work, we adapt two existing variational quantum algorithms, which we refer to as \textit{Parametric Gate Encoding (PGE)} \cite{new_nisq} and \textit{Ancilla Basis Encoding (ABE)} \cite{Tan2021qubitefficient}, to address the task of image segmentation formulated as a combinatorial optimization problem \cite{venkatesh2023qseg}.
\textit{PGE} encodes the segmentation solution in the parameterized gates exerted in the quantum ansatz, while \textit{ABE} \cite{Tan2021qubitefficient} uses the probability distribution of the qubits' basis states for encoding the solution \cite{new_nisq}.
Furthermore, we propose an enhancement of \textit{ABE}, termed \textit{Ancilla Cost Encoding (ACE)} that uses a problem-specific cost function and demonstrates an improved performance.
Importantly, these algorithms require exponentially fewer qubits than QAOA and its variants, scaling logarithmically with the number of image pixels.
This represents a significant improvement in scalability for image segmentation tasks using VQAs.
We begin by reformulating the image segmentation problem as a graph partition problem, subsequently framed as a QUBO.
% The efficacy of these algorithms is rigorously benchmarked against the QAOA, a well-known quantum algorithm for solving QUBO problems, to assess performance and scalability.
The experiments are focused on analyzing the solution quality and the efficiency of various classical optimizers for these algorithms.
% We also provide a theoretical analysis of the quantum resource requirements. 
We also provide a theoretical analysis of essential quantum resource requirements as a function of the number of pixels of the input image.
% —focusing on qubits, gates, and circuit depth—and an empirical evaluation of synthetically generated images against various classical optimizers and circuit depths.
% Through this work, we not only demonstrate the practicality and efficiency of both approaches in image segmentation but also discuss the specific limitations of each method.
For a well-rounded study, we also discuss the specific limitations of each approach.
Our findings open new research directions for leveraging quantum computing in computer vision, particularly in the optimization of quantum resources for complex computational tasks.

\vspace*{-2pt}

\section{Related Works}
\vspace*{-4pt}

Supervised deep learning models have set benchmarks in image segmentation by leveraging precisely annotated data.
However, the collection of such labeled datasets faces hurdles, including significant annotation time and costs, the necessity of specialized knowledge, and issues scaling to vast datasets, not to mention the inconsistencies arising from variable interpretations by different annotators \cite{RFB15a}.
These limitations highlight the importance of unsupervised segmentation methods, especially valuable for leveraging extensive unlabeled data pools \cite{Wang_2023}.
Nonetheless, the computational demand of some unsupervised techniques, like graph-based segmentation \cite{yi2012image}, presents a significant limitation, making these tasks ideal targets for quantum computing to promise more efficient solutions.
% Among these, graph-based segmentation emerges as a notable technique. It conceptualizes segmentation as dividing a graph, where each node is a pixel or region, connected by edges weighted according to pixel differences.
% This strategy is adept at incorporating both the edge and region data to identify meaningful clusters, offering optimal segmentation outcomes in certain scenarios \cite{yi2012image, 10.1117/12.919743}.

% The intersection of quantum computing and computer vision has sparked a range of innovative approaches to complex vision tasks, notably in the domain of image segmentation. This section reviews prior works across different quantum computing paradigms, highlighting their contributions and limitations in the context of image analysis.

A wide range of research has explored the use of quantum algorithms for computer vision tasks, primarily focusing on image classification \cite{neven2008image, neven2008training, neven2009nips, grant2018hierarchical} and matching \cite{Jiang2016, bhatia2023CCuantuMM}. Additionally, quantum-inspired classical techniques have been applied to segmentation \cite{aytekin2013quantum} and edge-detection \cite{yuan2013quantum, fu2009new}, blending quantum concepts with traditional computing infrastructure.
% The application of the Inverse Quantum Fourier Transform (I-QFT) \cite{akinola2023inverse} represents a theoretically promising yet practically challenging approach, requiring a fault-tolerant quantum environment. Similarly, quantum circuit-based thresholding methods \cite{Caraiman2015} demonstrate the theoretical potential of quantum computing for segmentation, though they often fall short of surpassing the efficiency of classical systems in less demanding scenarios \cite{otsu1979threshold, 10.1117/1.1631315}.
% Quantum Annealers and Optimization: Quantum annealers have been utilized to tackle the image segmentation problem formulated as a constrained optimization task, later converted into a QUBO problem \cite{presles2023synthetic, venkatesh2023qseg}. While this approach shows potential in scalability, its applicability is limited by the quantum annealers' current capabilities and the structural constraints of the problem's graph representation.
% Hybrid Quantum-Classical Approaches: The integration of quantum and classical computing techniques has been explored, with quantum annealers aiding in feature extraction followed by classical segmentation processing \cite{9323504}. Despite its innovative approach, this methodology faces scalability issues when applied to large datasets.
% Variational Quantum Circuits for Graph-Based Segmentation: Although variational quantum circuits offer a novel solution, their practicality is hindered by the extensive qubit requirements for high-resolution images, posing a significant scalability challenge.
In particular, quantum annealers and hybrid quantum-classical approaches have demonstrated promise for image segmentation, converting these challenges into a QUBO problem \cite{tse2018graph, presles2023synthetic, venkatesh2023qseg}.
Despite their innovative methods, they face limitations in scalability and are constrained by the current capabilities of quantum annealers for handling high-resolution images.
% QAOA has also been explored for graph-based image segmentation but is not scalable due to limited  \cite{tse2018graph}.

% In this context, our study introduces a novel approach by harnessing the capabilities of Variational Quantum Algorithms (VQAs), which are particularly well-suited for NISQ devices. This strategy seeks to transcend the scalability and resource constraints observed in existing quantum and hybrid methodologies. VQAs, and notably QAOA, are identified as more accessible and feasible options for the current quantum hardware, facilitating an efficient exploration of solution spaces for complex challenges such as image segmentation. Our research aims to fill the noted research gap by optimizing VQAs for enhanced quantum resource efficiency and scalability, targeting the improvement of computer vision techniques through quantum computing advancements.

In contrast, our work focuses on leveraging the versatility of VQAs, particularly suited for NISQ devices, to overcome the scalability limitations in existing quantum and hybrid approaches. Unlike methods requiring extensive quantum resources or fault tolerance, we adapt established encoding strategies to harness the potential of quantum circuits in enhancing computer vision techniques, specifically image segmentation, through efficient quantum resource utilization.

% \vspace*{-8pt}
\section{Preliminaries}
\vspace*{-4pt}
% \subsection{Graph-based Image Segmentation}
The segmentation process involves representing an input image as a lattice graph with nodes corresponding to pixels and edges representing pixel similarity.
The approach is particularly interesting as it can capture both spatial and spectral information in the image \cite{yi2012image}.
Segmentation seeks to partition this graph's vertices into mutually exclusive subsets, such that the total weight of the edges connecting vertices from one subset to another is minimized, defined as:
% Segmentation seeks to partition this graph into mutually exclusive subsets, such that the sum of the weights of the edges between the , defined as:
\begin{equation} \label{eqn: mincut}
    \textsc{MinimumCut}(G) = \arg \min_{A,\overline{A}} \sum_{i \in A, j \in \overline{A}} w(v_{i},v_{j})
\end{equation}
where $G$ denotes the graph and $w(v_{i},v_{j})$ is the weight of the edge connecting the nodes $v_{i}$ and $v_{j}$.
Given the NP-hard nature of the min-cut problem for arbitrary weights, classical solutions become computationally challenging, motivating the reformulation into a QUBO problem to exploit the power of quantum computers \cite{gcs-q}.

\paragraph*{Example}

Let's consider an image of size $2 \times 2$, which is converted to a grid graph with a one-to-one mapping of the pixels to the vertices.
The edge weights are assigned as the similarity measure between the neighboring pixels.
Finding the minimum cut in this grid graph will eventually give the segmentation of the image.

\vspace*{-10pt}

\begin{figure}[htbp]
\centering
\centerline{\includegraphics[width=\columnwidth]{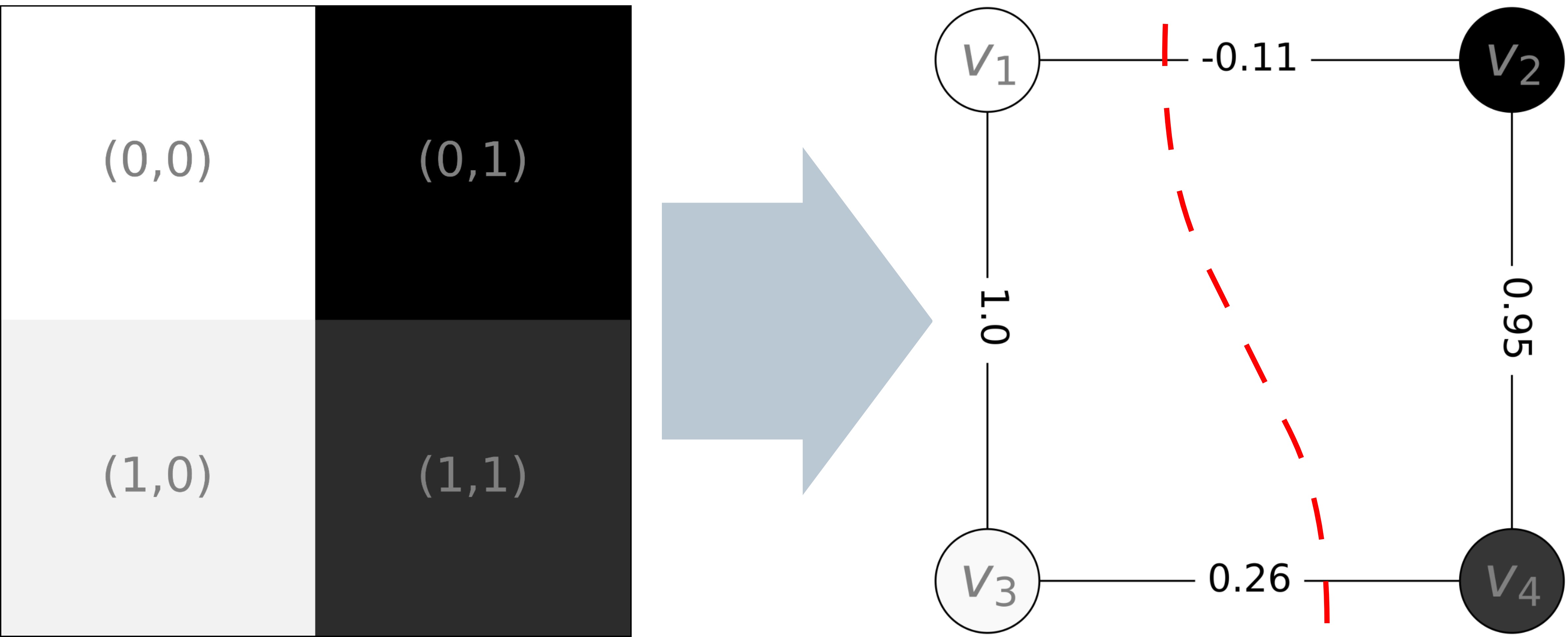}}
% \caption{Converting the pixel values into edge weights in the grid graph. The graph structure and the edge weights capture both spatial and spectral information.}
\vspace*{-6pt}
\caption{Transformation of a $2\times 2$ pixel image into a grid graph, where edge weights indicate pixel similarity. The red dashed line depicts the edges that are cut to partition the graph.}
\vspace*{-8pt}
\label{fig: example}
\end{figure}

% A near-optimal solution might be tolerable for the downstream computer vision tasks.
% Therefore, we focus on leveraging the capabilities of the existing NISQ technology.
Focusing on the capabilities of existing NISQ technology, we target near-optimal solutions as they are often tolerable for downstream computer vision tasks.
% For $n$ vertices, a binary encoding of each vertex of $G$ enables to represent a possible solution by a binary vector $\in \{0,1\}^n$, which allows to represent the vertices as two disjoint subsets and the edges between the two subsets are considered to be cut.
With $n$ vertices, the binary encoding of each vertex in graph $G$ facilitates the representation of potential solutions as vector $\in \{0,1\}^n$ delineating the vertices into two disjoint subsets, with edges intersecting these subsets identified as cuts.
Such an encoding allows to reformulate the minimum cut as a QUBO without loss of generality as\cite{Lucas2014, glover2018tutorial}: 
\vspace*{-5pt}
% A canonical form of the minimum cut is :
\begin{equation} \label{eqn: mincut canonical}
\begin{aligned}
    \vec{x}^* &= \arg \min_{\vec{x}} \sum_{1 \leq i < j \leq n} x_{v_{i}} (1-x_{v_{j}}) w(v_{i},v_{j}) \\
    % &= \arg \min_{\vec{x}}  \sum_{i=1}^{n} q_{i}x_{v_{i}} + \sum_{1 \leq i < j \leq n} q_{ij} x_{v_{i}} x_{v_{j}} \\
    &= \arg \min_{\vec{x}} \ {\vec{x}}^T Q {\vec{x}}
\end{aligned}
\end{equation}
\vspace*{-8pt}

where $\vec{x}^*$ represents the optimal solution for the minimum cut of $G$. 
% Expanding and simplifying the first expression
% obtains quadratic terms in $x_{v_{i}}$ where $x \in \{1,2,...,n\}$.
% allows coefficients of the linear terms as diagonal elements and the coefficients of the quadratic terms as the non-diagonal elements of a matrix $Q$.
% substituting the edge weights $w(v_{i},v_{j})$ and simplifying yields a quadratic expression in binary variables $x_{v_{i}}$ for $i \in \{1,2,...,n\}$.
The construction of the $Q \in \mathcal{R}^{n\times n}$ matrix is derived from the coefficients of terms in the quadratic expression of the binary variables $x_{v_{i}}$ for $i \in \{1,2,...,n\}$ obtained by substituting the edge weights $w(v_{i},v_{j})$ and simplifying.

QAOA, a special case of the Variational Quantum Eigensolver (VQE) \cite{10097179}, is designed for QUBO problems, implementing a parameterized quantum circuit to approximate the problem's ground state.
The binary encoding strategy employed for solving the image segmentation problem using QAOA leads to scalability challenges of representing high-resolution images with individual qubits for each pixel.
% The ansatz in the circuit of QAOA is problem-inspired, while the methods we discuss employ problem-agnostic ansatz.
In contrast to the problem-specific ansatz of QAOA, the methods we explore use a problem-agnostic ansatz.
This gives rise to a need for efficient encoding strategies within the VQA framework to address image segmentation, requiring fewer qubits and offering scalable solutions for practical applications.

\section{Methods}

In this section, we will explain the two methods and describe a new approach to obtain the segmentation of an image employing innovative encoding strategies for solving the QUBO problem. % in Eq. \ref{eqn: qubo cost function}.
% that are alternatives of QAOA which use exponentially fewer qubits.

% \subsection{Minimal Encoding \cite{Tan2021qubitefficient}}
\subsection[Parametric Gate Encoding (PGE)]{Parametric Gate Encoding (PGE)}

% This method exploits the natural correspondence between combinatorial optimization and quantum mechanics, leveraging the superposition and interference properties of quantum states to explore the solution space more efficiently than classical counterparts.
% The quantum algorithmic framework exploits the natural correspondence between combinatorial optimization and quantum mechanics for finding the minimum cut in a graph.
% \cite{chatterjee2023solving}.

% Here, we discuss a VQA for image segmentation that again requires exponentially fewer qubits than QAOA but uses a different encoding strategy.
% Given an image, the construction of the grid graph is similar to the previously discussed approach, but a more naive graph-theoretical concept is applied and the reformulation to a QUBO problem for finding the minimum cut can be omitted using the \textit{New NISQ Algorithm} \cite{new_nisq}.

% The graph \( G \) is encoded into a quantum-compatible representation via its Laplacian matrix \( L_G \), which is constructed as the difference between the degree matrix \( D_G \), a diagonal matrix with vertex degrees on the diagonal, and the adjacency matrix \( A_G \), which holds the edge weights:

Given a grid graph $G$ constructed from the input image, the Laplacian matrix $L_G$ provides a compact representation of the graph's structure, capturing information about the connectivity and degree of each vertex, and is given by $L_G = D_G - A_G$, where $D_G$ is the degree matrix, a diagonal matrix with vertex degrees $d(v_i)$ for $v_i$ on the diagonal, and $A_G$ is the adjacency matrix, a square matrix where the element $A_G[i,j]$ is the weight of the edge between the vertices $v_i$ and $v_j$.
If the dimension of the matrix $L_G$ is not a power of $2$, without loss of generality, we pad the matrix with zeroes to the nearest power of $2$ to ensure that the matrix dimensions are compatible with the quantum register, as quantum states are represented in a Hilbert space of dimension $n$ for $\log_2(n)$ qubits.
Since the Laplacian is a symmetric matrix consisting of real-valued edge weights for undirected weighted graphs, it is also Hermitian and thus serves as a suitable quantum observable \cite{new_nisq}.

% **2. Hermitian Matrix Formulation**:
% The Laplacian matrix \( L_G \), being symmetric, is extended to a Hermitian matrix \( H \) by padding with additional rows and columns to match a dimension that is a power of two, which facilitates quantum state representation:

% \[ H = \text{pad}(L_G) \]

% **3. Observable Decomposition**:

% \[ H_{\text{decomposition}} = \text{Decompose}(H) \]

% **4. Quantum Circuit Construction**:
% The \textit{PGE} approach \cite{new_nisq} involves constructing a parametrized quantum circuit is constructed with $n^\prime = \lceil \log_2(n) \rceil$ qubits and initially applied with Hadamard gates for equal superposition.
% As the ansatz, a diagonal gate is applied which can be expressed as \cite{new_nisq}:
The \textit{PGE} method constructs a parameterized quantum circuit with $n^\prime = \lceil \log_2(n) \rceil$ qubits, initializing with Hadamard gates for equal superposition and applying a diagonal gate as the ansatz which can be expressed as \cite{new_nisq}:
\begin{equation} \label{eqn: new nisq ansatz}
\resizebox{0.485\textwidth}{!}{
    $\hat{U}(\vec{\Theta}) = \begin{pmatrix}
        \exp(i\pi f(\theta_1)) & 0 & \cdots & 0 \\
        0 & \exp(i\pi f(\theta_2)) & \cdots & 0 \\
        \vdots & \vdots & \ddots & \vdots \\
        0 & 0 & \cdots & \exp(i\pi f(\theta_{2^{n^\prime}}))
    \end{pmatrix}$
}
\end{equation}

% \resizebox{0.489\textwidth}{!}{

% $$\resizebox{0.485\textwidth}{!}{
%     $\hat{U}(\vec{\Theta}) = \begin{pmatrix}
%         e^{i\pi f(\theta_1))} & 0 & \cdots & 0 \\
%         0 & e^{i\pi f(\theta_2))} & \cdots & 0 \\
%         \vdots & \vdots & \ddots & \vdots \\
%         0 & 0 & \cdots & e^{i\pi f(\theta_{N}))}
%     \end{pmatrix}$
% }$$

% \vspace*{1pt}

where $\vec{\Theta} = (\theta_1, \theta_2, ...  \theta_{2^{n^\prime}})$, and $\theta_i \in [0, 2\pi]$ for $i \in \{1,2,...,2^{n^\prime}\}$ and $f$ is piecewise function defined as:

\vspace*{-10pt}

\begin{equation} \label{eqn: new nisq encoding}
f(\theta_i) = x_{v_i} = \begin{cases}
        0 &  0 \leq \theta < \pi\\
        1 &  \pi \leq \theta < 2\pi \\
      \end{cases}
\forall i \in \{1,2,...,2^{n^\prime}\}
\end{equation}

% $$
% f(\theta_i) = x_{v_i} = \begin{cases}
%         0 &  0 \leq \theta < \pi\\
%         1 &  \pi \leq \theta < 2\pi \\
%       \end{cases}
% \forall i \in \{1,2,...,N\}
% $$

with $n^\prime-2$ two-qubit CNOT gates and $n^\prime$ single-qubit parametric gates, where each parametric gate in the variational ansatz encodes a binary decision for the segmentation mask.
With $L_G$ as the observable, the measurement operation is performed, and the energy of the system is evaluated as:
% \vspace*{-4pt}
\begin{equation} \label{eqn: new nisq cost}
    C(\vec{\Theta}) = \frac{2^{n^\prime}}{2}\langle\psi(\vec{\Theta})|L_G|\psi(\vec{\Theta})\rangle
\end{equation}

Since $L_G$ is also a Hermitian matrix, it is decomposed into a linear combination of tensor products of Pauli matrices, enabling the evaluation of the quantum observable corresponding to the graph structure.
The parameters $\vec{\Theta}$ are adjusted iteratively using a classical optimizer to minimize the cost function (Eq. \ref{eqn: new nisq cost}).
The optimal parameters $\vec{\Theta} = \{\theta_1, \theta_2, ...  \theta_{2^{n^\prime}}\}$ are decoded to binary values (Eq. \ref{eqn: new nisq encoding}), obtaining the minimum cut and eventually the binary segmentation mask of the image.

\begin{figure*}[htbp]
  \centering
  \includegraphics[width=\textwidth]{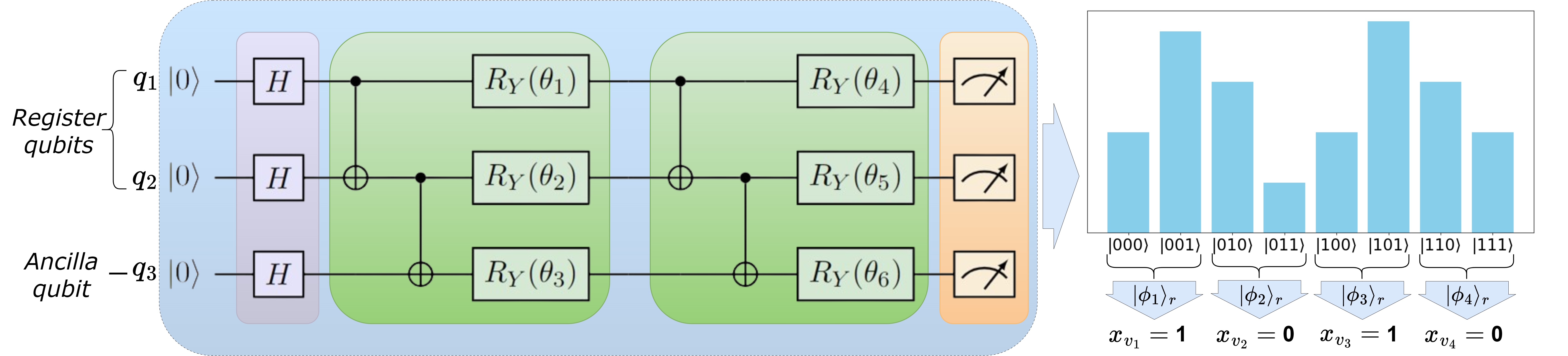}
  \vspace*{-16pt}
  \caption{Circuit schematic of \textit{ABE}/\textit{ACE} for finding the segmentation mask of the example image in Fig. \ref{fig: example}. illustrating the encoding of the solution of a QUBO problem of size $4$ into a quantum state facilitated by 3 qubits: $q_1, q_2$ are the register qubits and $q_3$ is the ancilla qubit.
  The circuit initialization employs Hadamard gates to induce superposition across all qubits, followed by two layers of hardware-efficient ansatz with entanglement and rotation gates parameterized by $\vec{\Theta}$.
  Executing and measuring the circuit obtains a probability distribution over the basis states. For the basis states whose register qubits are the same encodes the value of the binary variable. For example, with the register qubit basis states $|\phi_1\rangle_r = |00\rangle$, we obtain the value $x_{v_1} = 1$ as the probability of the ancilla qubit in state $|1\rangle > |0\rangle$ (Eq. \ref{eqn: minimal decoding}).}
  \vspace*{-16pt}
  \label{fig: example abe ace}
\end{figure*}

\subsection[Ancilla Basis Encoding (ABE)]{Ancilla Basis Encoding (ABE)}

% The ABE strategy enables using VQAs to solve the QUBO problem in which the number of qubits in the quantum circuit also scales logarithmically to the size of QUBO which is originally referred to as minimal encoding \cite{Tan2021qubitefficient}.

% \paragraph*{Encoding Strategy}

For an image with $n$ pixels to be segmented, the QUBO problem formulated will also have $n$ variables.
The \textit{ABE} \cite{Tan2021qubitefficient} strategy uses only $\log_2(n) + 1$ number of qubits, where the $\log(n)$ qubits are called the register qubits and the additional qubit is termed as the ancilla qubit.
Using a hardware efficient ansatz where each layers consists of $\log_2(n)$ CNOT gates between adjacent qubits and $\log_2(n) + 1$ parametric $R_y(\theta)$ gates for each qubit, the quantum state is represented as:
\vspace*{-4pt}
\begin{equation} \label{eqn: minimal encoding circuit}
|\psi(\vec{\Theta})\rangle = \sum_{i=1}^{n} \beta_i(\vec{\Theta}) (a_i(\vec{\Theta}) |0\rangle_a + b_i(\vec{\Theta}) |1\rangle_a) \otimes |\phi_i\rangle_r
\end{equation}
\vspace*{-8pt}

where $ |\phi_i\rangle_r $ are the computational basis states of the register qubits, $ |0\rangle_a $ and $ |1\rangle_a $ are the states of the ancilla qubit, $ a_i $ and $ b_i $ are the amplitudes for the ancilla qubit, and $ \beta_i(\vec{\Theta}) $ are coefficients dependent on variational parameters $ \vec{\Theta} $.
The expectation measurements obtain a probability distribution over the basis states of the qubits in the circuit and the solution is decoded as:
% The measurement operation, which is typically performed several times, gives a probability distribution over the basis states of the qubits in the circuit and the solution is decoded as:
\vspace*{-8pt}
\begin{equation} \label{eqn: minimal decoding}
x_{v_i} = \begin{cases} 
        0 & |a_i|^2>|b_i|^2 \\
        1 & otherwise \\
      \end{cases}
\vspace*{-6pt}
\end{equation}
where $ |a_i|^2 + |b_i|^2 = 1 $.

The probability of finding the optimal solution increases as the number of measurements tends to $\infty$ \cite{Tan2021qubitefficient}.
% Although this may seem to be impractical, this is common with VQAs, and even the eminent QAOA claims to find the solution as the number of layers tends to $\infty$.
While it may appear impractical, for relatively small quantum circuits, the number of measurement operations only needs to be large enough to adequately approximate the optimal probability distribution; it does not necessarily need to be $\infty$.
Additionally, well-established VQAs like QAOA are guaranteed to find the optimal solution only as the depth of the corresponding quantum circuit tends to infinity.
According to the ABE approach, the quantum circuit is typically initiated by a layer of Hadamard gates to generate a uniform superposition, followed by a hardware-efficient ansatz applied to evolve the quantum state.
% The variational ansatz $\hat{U}(\vec{\Theta})$ consists of one or more layers and each layer corresponds to a sequence of CNOT entanglement gates and $R_y(\theta)$ parametric gates.
% Here, the $R_y$ gate is the only parameterized gate.
% The cost function in a variational quantum algorithm plays a pivotal role in mapping a QUBO problem to a quantum framework. 
% Here, the cost function for optimization is defined as:
For the parameters $\vec{\Theta}$, the cost is given by:
\vspace*{-4pt}
\begin{equation} \label{eqn: minimal encoding cost}
C(\vec{\Theta}) = \sum_{i,j=1}^{n} Q_{ij} \frac{\langle \hat{P}^1_i \rangle_{\vec{\Theta}} \langle \hat{P}^1_j \rangle_{\vec{\Theta}}}{\langle \hat{P}_i \rangle_{\vec{\Theta}} \langle \hat{P}_j \rangle_{\vec{\Theta}}} (1 - \delta_{ij}) + \sum_{i=1}^{n} Q_{ii} \frac{\langle \hat{P}^1_i \rangle_{\vec{\Theta}}}{\langle \hat{P}_i \rangle_{\vec{\Theta}}}
\end{equation}
\vspace*{-8pt}

where $ \hat{P}_i $ are projectors over the basis states of the register qubits $ |\phi_i\rangle_r $, irrespective of the state of the ancilla qubit, $\delta_{ij} =1$ when $i=j$ and $Q$ is the matrix representing the QUBO problem.
% Here, the aim is to minimize the objective function characterized by the matrix $Q$.
The objective is to find a binary vector $\vec{x}^*$ that minimizes $\vec{x}^\top Q \vec{x}$.
% The QUBO problem's objective function is transcribed into the cost function of the quantum domain such that the expectation value reflects the original objective function.
The expectation values $\langle \hat{P}_i \rangle$ embody the probabilities of observing the qubit states that encode the binary variables $x_{v_i}$ of the QUBO solution vector $\vec{x}$.
% These probabilities, weighted by the QUBO matrix $Q$, are then used to compute the expectation value of the cost function.
The projectors $P_i$ and $\hat{P}^1_i$ are operators that map the quantum state onto the respective basis states of the register qubits, with $P_i$ aligning with the state that encodes $x_{v_i} = 1$. $\hat{P}^1_i$ modifies this to include the state of the ancilla qubit \cite{Tan2021qubitefficient}.
Finally, an optimizer running on a classical computer optimizes the variational parameters $\vec{\Theta}$ to minimize  $C(\vec{\Theta})$ (Eq. \ref{eqn: minimal encoding cost}).

% The objective to find a binary vector $\vec{x}$ that minimizes $\vec{x}^\top Q \vec{x}$ is incorporated in finding the minimum of the cost function in Eq. \ref{eqn: minimal encoding cost} and has been comprehensively proven \cite{Tan2021qubitefficient}.

% In quantum optimization, the cost function evaluates the proximity of a given quantum state to the problem's solution.

% \paragraph*{Optimization}

% In summary, one step of optimization involves the preparation of the quantum state $|\psi(\vec{\Theta})\rangle$, its measurement in the computational basis, and the update of $\vec{\Theta}$ using a classical optimizer until convergence.

\subsection{Adaptive Cost Encoding (ACE)}

% In ABE, the cost function in Eq. \ref{eqn: minimal encoding cost} has a non-linear relation with the cost of the min-cut.
% Although the authors argue that the decrease in the ABE cost does not correspond to a decrease in the QUBO problem's cost.
% Thus, 
We propose a modified cost function using the same encoding strategy as \textit{ABE}.
During each iteration of the optimization, once we measure the circuit multiple times, we get a probability distribution that can be decoded into a binary vector $\vec{x}$ that represents a possible solution.
Instead of using the cost function in Eq. \ref{eqn: minimal encoding cost}, we can make use of the min-cut problem's cost function i.e., 

\vspace*{-8pt}

\begin{equation} \label{eqn: mincut cost function}
    % x^* = \arg \min_x \sum_{v_i \in A, v_j \in \overline{A}} x_{v_{i}} (1-x_{v_{j}}) w(v_{i},v_{j})
    C(\vec{x}) = \sum_{1 \leq i < j \leq n} x_{v_{i}} (1-x_{v_{j}}) w(v_{i},v_{j})
\vspace*{-6pt}
\end{equation}
% $\text{where } x = x_{v_{1}}, x_{v_{2}},...,x_{v_{n}} \text{ such that } x_{v_{i}} \in \{0,1\} \text{,and } w(v_{i},v_{j}) \text{ is the edge weight between} v_{i} \text{ and } v_{j}$.
\text{where } $\vec{x} = x_{v_{1}}, x_{v_{2}},...,x_{v_{n}}$  such that $x_{v_{i}} \in \{0,1\}$ given by Eq. \ref{eqn: minimal decoding}, and $w(v_{i},v_{j})\ \forall\ i,j \in \{1,2,...,n\}$ is the edge weight between $v_{i}$ and $v_{j}$.
% The min-cut cost function in Eq. \ref{eqn: mincut cost function} in our usecase.
In case of \textit{ABE}, given two sets of parameters, say $\vec{\Theta}_1$ and $\vec{\Theta}_2$, obtaining the binary vectors $\vec{x}_1$ and $\vec{x}_2$ respectively (Eq. \ref{eqn: minimal encoding circuit} and \ref{eqn: minimal decoding}), if $C(\vec{\Theta}_1) < C(\vec{\Theta}_2)$ (Eq. \ref{eqn: minimal encoding cost}) does not always guarantee that $C(\vec{x_1}) < C(\vec{x_2})$ (Eq. \ref{eqn: mincut cost function}) which implies that the optimal circuit parameters may not correspond to the minimum cut.
Whereas, \textit{ACE} guarantees that the optimization process obtains the solution of the original problem, i.e., the optimal circuit parameters $\vec{\Theta}$ implies the corresponding $\vec{x}$ minimizes the minimum cut cost function (Eq. \ref{eqn: mincut cost function}).
Moreover, the QUBO formulation of the original problem also becomes obsolete as the cost function directly involves the weights of the edges in the graph and not the QUBO coefficients.
In the following section, through our experiments, we demonstrate ACE to have improved trainability, a strong tendency to find better solutions and a more consistent behavior using different optimizers with our proposed problem-specific cost function compared to \textit{ABE}.

Furthermore, the adaptive cost function can be extended to other combinatorial optimization problems where the problem formulation expresses the solution as a binary vector and uses the original cost function.
This approach can be crucial for tasks where the optimization problem includes constraints and reformulating the problem as a QUBO involves integrating the constraints as penalty terms into the cost function with an appropriate penalty coefficient, eventually not depicting the original problem \cite{bilp-q}.
Even an optimal penalty coefficient should be set, while a very low value can lead to the consideration of an invalid binary vector as a solution, or a high value can cause any valid solution to be considered optimal.
Whereas, \textit{ACE} allows finding the optimal solution of the original problem without modifying the cost function or the constraints, by setting a high or low cost for binary vectors that do not obey the constraints depending on whether the problem is minimization or maximization respectively.

\begin{figure*}[htbp]
  \centering
  \includegraphics[width=\textwidth]{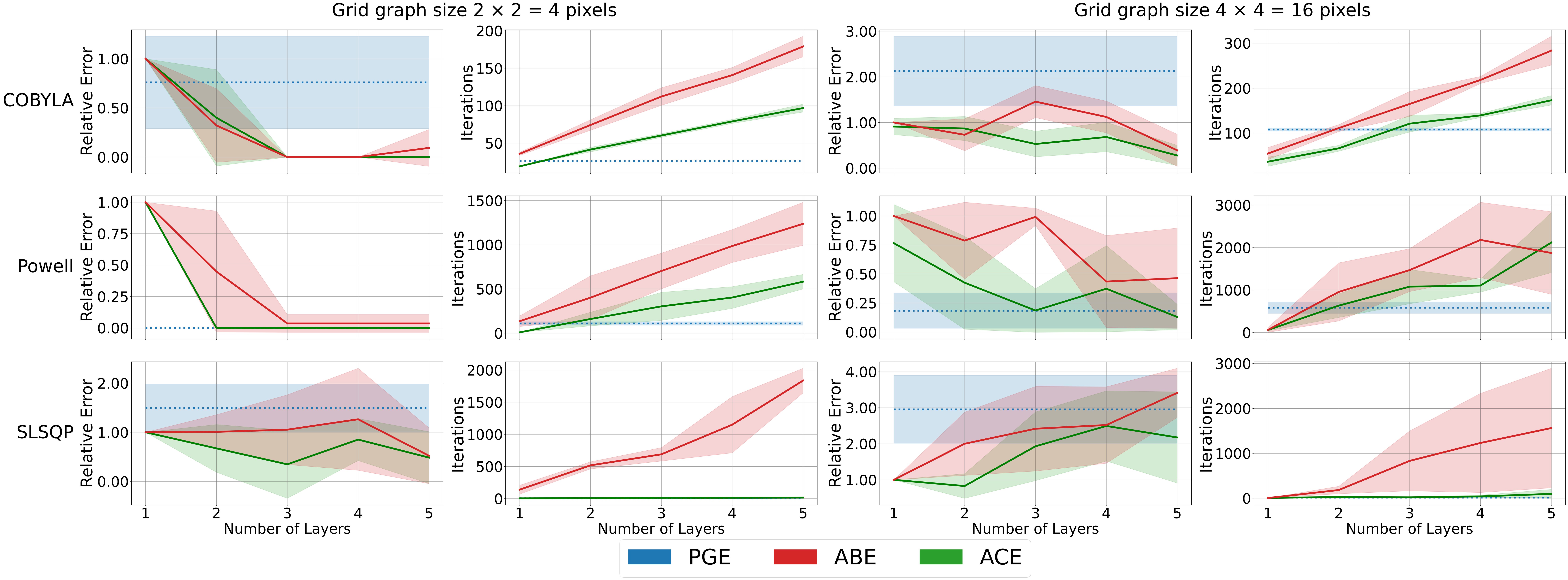}
  \vspace*{-16pt}
  \caption{The figure illustrates the performance of \textit{PGE}, \textit{ABE} and \textit{ACE} for finding the minimum cut on grid graphs of size $2 \times 2$ and $4 \times 4$ with the layers in the ansatz of the quantum circuit ranging from $1$ to $5$ in terms of the solution quality and the efficiency of the optimization process. The plots show the average and standard deviation of the relative errors in the obtained solution's cost as a measure of quality and the number of iterations to quantify the trainability using COBYLA, Powell, and SLSQP optimizers.}
  \vspace*{-16pt}
  \label{fig: exp results 4 16}
\end{figure*}

\paragraph*{Example}

Considering again the $2 \times 2$ image from the previously discussed example (Fig. \ref{fig: example}).
% Note that the QUBO formulation of the minimum cut does not introduce any overhead depending on the number of binary variables compared to the input size in the original problem, as is the case with other existing formulations \cite{venkatesh2023qseg}.

For \textit{PGE}, the Laplacian $L_G$ is of size $4 \times 4$ and the encoding scheme uses a quantum circuit with just $2$ qubits.
The circuit consists of an initial layer of Hadamard gates, followed by the ansatz $\hat{U}(\vec{\Theta})$ (Eq. \ref{eqn: new nisq ansatz}) which is a diagonal gate of size $4 \times 4$.
With $L_G$ as the observable, optimizing the cost function and obtaining the optimal circuit parameters $\theta^*_1, \theta^*_2, \theta^*_3, \theta^*_4$, we can decode the binary mask $x^*_{v_1}, x^*_{v_2}, x^*_{v_3}, x^*_{v_4}$ (Eq. \ref{eqn: new nisq encoding}).
While this approach is efficient in terms of the number of qubits, the main drawback is the number of parameters which implies an overhead in the classical optimization process.

Figure \ref{fig: example abe ace} represents the quantum circuit and the encoding strategy for solving this example.
In the cost function (Eq. \ref{eqn: minimal encoding cost}) of \textit{ABE}, the term $\langle \hat{P}_i \rangle_{\vec{\Theta}}$ is the probability of the basis state $(|0\rangle_a + |1\rangle_a) \otimes |\phi_i\rangle_r$, and the term $\langle \hat{P}^1_i \rangle_{\vec{\Theta}}$ is the probability of the state $|1\rangle_a \otimes |\phi_i\rangle_r$, where $|0\rangle_a$ and $|1\rangle_a$ are the basis states of the ancilla qubit.
Updating the parameters, a unitary evolution is performed over the state $|\psi_0\rangle$ which produces a new state $|\psi_1(\vec{\Theta})\rangle = \hat{U}_1(\vec{\Theta})|\psi_0\rangle$.
The optimization iteratively adjusts $\vec{\Theta}$, and each iteration involves running the quantum circuit and evaluating the cost function.
A classical optimizer is used to find the optimal parameters $\vec{\Theta}_{\text{opt}}$ that minimize the cost function $C(\vec{\Theta})$.
Decoding the final quantum state $|\psi_1(\vec{\Theta}_{\text{opt}})\rangle$, we obtain a binary vector $\vec{x}$ which represents the solution of the QUBO problem, which is the binary segmentation mask of the given image.
Whereas in the case of \textit{ACE}, at each iteration of the optimizer, the measurement is decoded to get the binary vector $\vec{x}$, and the value of the cost function in Eq. \ref{eqn: mincut cost function} is minimized by optimizing the circuit parameters $\vec{\Theta}$.

\section{Experiments}

\subsection{Experimental Settings}

\textit{PGE}, \textit{ABE}, \textit{ACE} algorithms are implemented in Qiskit framework, and the experiments are executed on a classical computer using Qasm simulator.
Inspired from the existing literature \cite{venkatesh2023qseg}, the synthetic dataset consist of square grid graphs by sampling edge weights from a uniform distribution in the range [-1,1], and the experiments were executed for sizes $2 \times 2$ i.e., 4 pixels, and $4 \times 4$ i.e., 16 pixels, as all the approaches discussed previously behave similarly for images of sizes from $\log_2(n+1)$ pixels up to size $n$.
For the reproducibility of results and fair comparison, we have executed the experiments using $5$ fixed seeds for generating the problem instances and the initial values for the parameters of the variations circuits.
% The available optimizer methods in Scipy and the genetic algorithm by DEAP have been tested and the results are presented for the best-performing ones.
% The local optimizers COBYLA, Powell, SLSQP in Scipy have been tested and the results are presented.
% In particular, \textit{ABE} and \textit{ACE} have been executed with increasing the number of layers, whereas the same does not apply for \textit{PGE}.
% The experiments for \textit{ABE} and \textit{ACE} have been executed for a maximum of $5$ layers as more layers imply that more parameters are included for optimization, causing significant overhead for the classical optimizer.
% As the solution quality increases as the number of measurements tends to $\infty$ \cite{Tan2021qubitefficient}, we executed our experiments with $65,536$ measurements at each iteration of the classical optimizer.
% As a classical benchmark, the results from a state-of-the-art solver \textit{Gurobi} have been considered for quality analysis.
% All software has been developed using \textit{Python 3.10}.
We evaluated the local optimizers from the Scipy library, detailing the results for COBYLA, Powell, and SLSQP.
For \textit{ABE} and \textit{ACE}, we increased the number of layers up to a maximum of $5$, allowing additional layers to introduce more parameters for optimization, causing significant overhead for the classical optimizer.
To ensure the solution quality improves with an increased number of measurements \cite{Tan2021qubitefficient}, we conducted our experiments with $65,536$ measurements for each optimization iteration.

% We benchmarked the solution quality and optimization efficiency of \textit{PGE}, \textit{ABE}, and \textit{ACE} against the results obtained from the state-of-the-art solver \textit{Gurobi}, renowned for its efficacy in solving combinatorial optimization problems. 
% To assess solution quality, we calculated the relative error of the minimum cut value found by \textit{PGE}, \textit{ABE}, and \textit{ACE} against the solution provided by \textit{Gurobi}, defined as $\text{Relative Error} = \frac{\left| C_{\text{VQA}} - C_{\text{gurobi}} \right|}{\left| C_{\text{gurobi}} \right|}$. We gauged the efficiency of the optimization process by the number of iterations required for the cost function, noting that the frequency of quantum circuit execution scales linearly with this metric. 
% Notably, the \textit{PGE} method, employs a constant-depth quantum circuit without any variance across the different number of layers in the ansatz of the quantum circuit. 
% All software implementations were developed in \textit{Python 3.10}.

We benchmarked the solution quality and optimization efficiency of \textit{PGE}, \textit{ABE}, and \textit{ACE} against the results from \textit{Gurobi}, known for its effectiveness in solving combinatorial optimization problems.
Solution quality was evaluated by calculating the relative error of the minimum cut value from the VQAs relative to \textit{Gurobi}, defined as $\text{Relative Error} = \frac{\left| C_{\text{VQA}} - C_{\text{gurobi}} \right|}{\left| C_{\text{gurobi}} \right|}$.
We measured optimization efficiency by the number of iterations needed for the cost function, which directly correlates with the frequency of quantum circuit executions.
\textit{PGE} uses a constant-depth quantum circuit and shows no variance across different layers of the ansatz. The software for these experiments was developed using \textit{Python 3.10}.

% We compare the performance of \textit{PGE}, \textit{ABE}, and \textit{ACE} in terms of the quality of the solution and the efficiency in the optimization process.
% As a benchmark for quality, we also solved the problem instances using an off-the-shelf solver \textit{Gurobi}.
% As \textit{Gurobi} is a state-of-the-art approach for solving combinatorial optimization problems, the relative error in the value of the minimum cut found by \textit{PGE} ($C_{\text{PGE}}$), \textit{ABE} ($C_{\text{ABE}}$) and \textit{ACE} ($C_{\text{ACE}}$) in comparison to that of \textit{Gurobi} ($C_{\text{gurobi}}$) is used as the metric for determining the quality of the solution, i.e.,
% $\text{Relative Error} = \frac{\left| C_{\text{ABE/ACE}} - C_{\text{gurobi}} \right|}{\left| C_{\text{gurobi}} \right|}$.
% The efficiency of the optimization process is estimated in terms of the number of iterations the cost function was invoked.
% The optimization iterations is an empirical metric for evaluating the time complexity as the number of times the quantum circuit is executed scales linearly to it.
% Furthermore, \textit{PGE} method employs a constant depth quantum circuit and there are no layers, thus shows no variance across different layers.

\subsection{Results}

\vspace*{-6pt}

Figure \ref{fig: exp results 4 16} illustrates the experiments for synthetically generated problems fixing $5$ seeds with layers $\in [1,5]$ using different classical optimizers.
We see that using our modified cost function in \textit{ACE} parameterized by the binary vector decoded from the measurement of the quantum circuit helps the optimizer decide upon the optimal solution much quicker than \textit{ABE} that uses a cost function parameterized directly by the probability distribution obtained from the measurement process.
Moreover, the training is more consistent for \textit{ACE} as we observe minimal variations in the number of iterations and the quality across $5$ seeds.
The number of iterations is consistently lower for \textit{ACE} as the different costs explored during the optimization are discrete as the minimum cut is a combinatorial optimization problem, lowering the possibilities to explore in contrast to \textit{ABE} where the optimization landscape is not only non-convex but also infinitely many values are explored.
However, certain optimizers, such as SLSQP, demonstrate limitations when faced with a higher number of layers, struggling with the increased overhead from optimizing a larger parameter set and thus proving unsuitable for scenarios demanding extensive parameter optimization.
% for SLSQP suggesting that SLSQP suffers from the overhead of optimizing the larger number of parameters and the optimizer is not a good fit for optimizing the circuit parameters.
Although, \textit{PGE} can be seen performing better with Powell optimizer, the number of parameters optimized is exponential to that of qubits.
Therefore, we extend our experiments using a more robust classical optimizer to prove the efficacy of \textit{ACE} in terms of the number of parameters to optimize.

% As the problem we are addressing is non-convex and non-differentiable, the classical optimizers implemented in the previous experiments that come under local optimization do not guarantee to find the global optimal solution
% % \footnote{https://docs.scipy.org/doc/scipy/reference/optimize.html}.
% % As the problem we are addressing is non-convex and non-differentiable, the classical optimizers implemented in the previous experiments that come under local optimization \textcolor{red}{[give a footnote to scipy]} cannot guarantee convergence.
% Although inefficient in terms of runtime, a more sophisticated method would be to use \textit{Differential Evolution} as the cost function is not smooth.
% This allows us to demonstrate that a single layer of \textit{ACE} also possesses the expressibility of representing the solution of the original problem, which has only a logarithmic number of parameters equal to the number of qubits.

Given the non-convex and non-differentiable nature of the problem, classical optimizers focused on local optimization may not always secure the global optimal solution.
Although inefficient in terms of runtime, using a metaheuristic optimizer like \textit{Differential Evolution} could be more effective as the cost function lacks smoothness. This approach helps illustrate that even a single layer of \textit{ACE}, with just a logarithmic number of parameters equal to the number of qubits, can capably represent the solution to the original problem.

% \vspace{-12pt}

\begin{table}[htbp]
\centering
\captionsetup{justification=justified,singlelinecheck=false}
% \resizebox{0.489\textwidth}{!}{
\resizebox{\columnwidth}{!}{
\begin{tabular}{|c|c|c|c|c|c|}
\hline
\textbf{Seed} & \textbf{Size} & \textbf{Iterations} & \textbf{Execution Time (s)} & \textbf{Obtained Value} & \textbf{Exact Value} \\
\hline
\multirow{2}{*}{111} & 4  & 3463  & 236.13 & -1.13 & -1.13 \\ %\cline{2-6} 
 & 16 & 14601 & 1320.92 & -1.54 & -1.54 \\
\hline
\multirow{2}{*}{222} & 4  & 4966  & 362.44 & -0.62 & -0.62 \\ %\cline{2-6} 
 & 16 & 75225 & 3557.49 & -3.97 & -3.97 \\
\hline
\multirow{2}{*}{333} & 4  & 5439  & 379.46 & -1.16 & -1.16 \\ %\cline{2-6} 
 & 16 & 40287 & 3477.47 & -3.57 & -3.57 \\
\hline
\multirow{2}{*}{444} & 4  & 9174  & 622.83 & -1.20 & -1.20 \\ %\cline{2-6} 
 & 16 & 75201 & 3797.97 & -2.19 & -2.19 \\
\hline
\multirow{2}{*}{555} & 4  & 3737    & 268.69 & -1.86 & -1.86 \\ %\cline{2-6} 
 & 16 & 75201 & 4202.49 & -2.56 & -2.56 \\
\hline
\end{tabular}
}
\vspace*{4pt}
% \caption{Results of \textit{ACE} with one layer using differential evolution optimizer. Specifies the problem generation seed for reproducibility, size to indicate the number of pixels, iterations denote the optimizer steps, and execution time is the runtime in seconds. Finally, the obtained value is the value of the minimum cut found by \textit{ACE}, compared to the exact value by a brute force method.}
\caption{Table details performance metrics for \textit{ACE} with a single layer, optimized using a differential evolution strategy. It includes reproducibility seeds, image sizes denoted by pixel count, total optimizer iterations, computational runtimes in seconds, and minimum cut values obtained by \textit{ACE} alongside those calculated through an exhaustive search for validation.}
\label{tab: exact results}
\vspace{-20pt}
\end{table}

The results in Table \ref{tab: exact results} reveal that \textit{ACE} can solve a combinatorial optimization problem consisting of $n$ variables by optimizing only $\log_2(n)+1$ parameters.
From a classical perspective, treating the quantum circuit as a black box, \textit{ACE} allows to reformulate an NP-Hard problem of size $n$ as a multivariate optimization problem of size $O(\log_2(n))$ leveraging the power of quantum computing.
% Furthermore, the quantum resources also scale logarithmically to the problem size.
In contrast, \textit{PGE} requires to optimize the same number of real-valued parameters and also uses more number of gates.

\vspace*{-4pt}

\subsection{Theoretical Analysis}

\vspace*{-4pt}

Here we perform an analysis on the scaling of \textit{QAOA}, \textit{PGE}, \textit{ABE} and \textit{ACE} based on the metrics typically considered for evaluating the scalability of VQAs.
In particular, we provide the complexities of the discussed NISQ algorithms for image segmentation in terms of qubit complexity which is a major bottleneck for present-day applicability, the number of two-qubit \textit{entanglement gates} is vital to understanding the error rates due to decoherence, the number of \textit{parametric gates} is crucial for assessing the overhead on the classical optimizer, and the circuit \textit{depth} which is the longest sequence of quantum gates applied from the input till the measurement operation which provides an estimate of circuit execution time.

% \begin{table}[ht]
% \centering
% % \footnotesize
% \resizebox{0.489\textwidth}{!}{
% \begin{tabular}{ | p{0.15\linewidth} | p{0.16\linewidth} | p{0.18\linewidth} | p{0.25\linewidth} | p{0.25\linewidth} | }
% \hline
% \textbf{Method} & \textbf{Qubit Complexity} & \textbf{Entanglement gates} & \textbf{Parametric gates} & \textbf{Circuit Depth} \\
% \hline
% \textbf{QAOA} & $O(n)$ & $O(n^2)$ & $O(L*n)$ & $O(L*n^2)$\\
% \hline
% \textbf{PGE} & $\log_2(n)$ & $n-1$ & $\log_2(n)$ & $n$\\
% \hline
% \textbf{ABE}/\textbf{ACE} & $\log_2(n) + 1$ & $L(\log_2(n))$ & $L(\log_2(n)+1)$ & $L(\log_2(n)+1)$\\
% \hline
% \end{tabular}
% }
% \caption{Scalability of \textit{QAOA, PGE, ABE, ACE} in terms of quantum resources}
% \label{tab: theoretical analysis}
% \end{table}

\begin{table}[htbp]
\centering
% \resizebox{0.489\textwidth}{!}{
\resizebox{\columnwidth}{!}{
% \begin{tabular}{ | p{0.15\linewidth} | p{0.16\linewidth} | p{0.21\linewidth} | p{0.22\linewidth} | p{0.25\linewidth} | }
\begin{tabular}{ | >{\centering\arraybackslash}m{0.15\linewidth} | >{\centering\arraybackslash}m{0.16\linewidth} | >{\centering\arraybackslash}m{0.21\linewidth} | >{\centering\arraybackslash}m{0.24\linewidth} | >{\centering\arraybackslash}m{0.24\linewidth} | }
% \begin{tabular}{|c|c|c|c|c|}
% \toprule
\hline
\textbf{Method} & \textbf{Qubit complexity} & \textbf{Entanglement gates} & \textbf{Parametric gates} & \textbf{Circuit depth} \\
\hline
\rule{0pt}{2.75ex}\textbf{QAOA} & $O(n)$ & $O(n^2)$ & $O(Ln)$ & $O(Ln^2)$\\
\rule{0pt}{3ex}\textbf{PGE} & $\log_2(n)$ & $n-1$ & $n$ & $n$\\
\rule{0pt}{3ex}\textbf{ABE}/\textbf{ACE} & $\log_2(n) + 1$ & $L(\log_2(n))$ & $L(\log_2(n)+1)$ & $L(\log_2(n)+1)$\\
\hline
% \bottomrule
\end{tabular}
}
\vspace{4pt}
% \captionsetup{justification=raggedleft, labelsep=newline}
\caption{Scalability analysis of VQAs for solving QUBO problems in terms of quantum resources.}
\label{tab: theoretical analysis}
\vspace{-12pt}
\end{table}
Table \ref{tab: theoretical analysis} provides a theoretical scalability analysis, where $n$ is the number of binary variables in QUBO and $L$ is the number of layers in the ansatz of the quantum circuit.
Acknowledging the qubit limitations in emerging quantum technologies, the scalability of the methods introduced here, excluding QAOA, is notably superior.
They potentially allow for the segmentation of a $1-$megapixel image, equivalent to $2^{20}$ pixels, utilizing merely $20$ qubits.

\section{Discussion and Conclusion}

% The methods applied for performing segmentation theoretically proclaim significant potential in scalability concerning the quantum resources contrasting the eminent VQAs like QAOA.
% For \textit{PGE}, there is also no overhead in terms of the circuit depth, and the ansatz employed is constant, fixing up on the number of vertices in the graph, irrespective of the number of edges.
% Specifically, the number of entanglement gates i.e., the two-qubit CNOT gates is lesser than the number of variables in the QUBO.
% However, the number of parameters is equivalent to the number of pixels which is exponential to the number of qubits in the circuit and the major bottleneck for the classical optimizer.
% As the number of real-valued circuit parameters to optimize is the same as the number of binary variables, a brute-force approach to find the minimum cut might be more efficient.
% Also, the processing power of a quantum computer is utilized mainly for evaluating the cost function, and the amount of resources spent turns out to be the same irrespective of the connectivity in the graph.
% The circuit remains the same for sparse and fully connected graphs, whereas the circuit in QAOA depends on the edges in the graph.
% Thus, a promising direction for \textit{PGE} is using more connections in the graph generated from the image that can lead to better consideration of the spatial information of each pixel with no extra overhead in classical computation.

The segmentation methods explored in this paper demonstrate substantial scalability in terms of quantum resources compared to well-known variational quantum algorithms like QAOA. 
For \textit{PGE}, there is no increase in circuit depth, and the ansatz remains constant regardless of the number of edges in the graph, only depending on the number of vertices.
This implies that while the number of entanglement gates, remains fewer than the number of variables in the QUBO, the overall resource efficiency is maintained.
The number of parameters, which corresponds directly to the number of pixels, represents the primary challenge for the classical optimizer due to its exponential relationship with the number of qubits. 
The consistency in circuit design across different graph connectivities suggests that \textit{PGE} could benefit significantly from more intricate connections in the image-derived graph. 
Such enhancements could potentially improve the spatial information consideration for each pixel, enriching segmentation quality without adding computational overhead. 

The methods \textit{ABE} and \textit{ACE} similarly display a significant reduction in quantum resource requirements for addressing NP-Hard segmentation problems, aligning with the conservation goals typical of near-term quantum technologies.
These methods achieve scalability primarily through their linear increase in parameters with the number of layers, which enhances their applicability without exponential increases in resource demands.
% Despite challenges related to the expressibility and completeness of solutions when measurements are restricted, \textit{ABE} and \textit{ACE} effectively mitigate these issues by judiciously increasing the depth of layers when necessary.
However, the number of circuit measurements scales exponentially with the number of qubits allowing us to measure the ancilla qubit state for all possible register basis states to obtain a complete binary vector solution.
Experimentally, \textit{ACE} has proven to achieve optimal solutions with increased layers and parameters, provided that a robust classical optimizer is used.
This highlights the potential for optimizing the synergy between quantum circuit design and classical optimization techniques, advancing applications of quantum technology.
Additionally, a deeper exploration of the optimization landscape could unveil methods allowing classical optimizers to find global optima more effectively.

% Similarly, the approaches \textit{ABE} and \textit{ACE} also mark a reduction in quantum resources for solving NP-Hard problems, with parameters scaling linearly with the number of layers.
% Although there is a challenge with expressibility when the number of measurements is limited, potentially leading to incomplete solutions, this limitation can be mitigated by increasing the layer depth.
% However, this does increase the number of parameters, influencing the efficiency of the optimization process.
% Despite this, \textit{ACE} has demonstrated, through experimental validation, that with a robust classical optimizer, it is possible to achieve optimal solutions efficiently.
% Additionally, a deeper analysis of the optimization landscape could potentially unveil specific characteristics that classical optimizers could leverage to more effectively locate global optima.

In summary, the techniques covered here demonstrate the potential of new encoding schemes, allowing vital applications like image segmentation to benefit from emerging quantum technologies, even with the current limitations of hardware from the NISQ era.
% These advancements open up promising avenues for future research into quantum-enhanced computer vision, underlining the critical role of innovative quantum computational approaches in overcoming traditional computational challenges.
These developments highlight the crucial role that novel quantum computational techniques play in overcoming classical computational challenges and open up promising directions for future research into quantum-enhanced computer vision.

\bibliographystyle{IEEEtran}
\bibliography{references}

\end{document}